\documentclass{ifacconf}

\usepackage{graphicx}      
\usepackage{natbib}        
\usepackage{url}
\usepackage{amssymb}

\usepackage{booktabs}

\begin{document}
\begin{frontmatter}
\title{Planetary Terrain Datasets and Benchmarks for Rover Path Planning}
\thanks[footnoteinfo]{
This work was partially supported by Kempestiftelserna (The Kempe Foundations), the Wallenberg AI, Autonomous Systems and Software Program (WASP) funded by the Knut and Alice Wallenberg Foundation, and the Swedish National Space Agency (SNSA).
}

\author[First]{Marvin Chanc\'an} 
\author[First]{Avijit Banerjee} 
\author[First]{George Nikolakopoulos}

\address[First]{Department of Computer Science, Electrical and Space Engineering (Robotics and AI Lab),
Lule\aa \: University of Technology,
Sweden
(e-mail: marvin.chancan@associated.ltu.se)}

\begin{abstract} 
Planetary rover exploration is attracting renewed interest with several upcoming space missions to the Moon and Mars. However, a substantial amount of data from prior missions remain underutilized for path planning and autonomous navigation research. As a result, there is a lack of space mission-based planetary datasets, standardized benchmarks, and evaluation protocols. In this paper, we take a step towards coordinating these three research directions in the context of planetary rover path planning. We propose the first two large planar benchmark datasets, MarsPlanBench and MoonPlanBench, derived from high-resolution digital terrain images of Mars and the Moon. In addition, we set up classical and learned path planning algorithms, in a unified framework, and evaluate them on our proposed datasets and on a popular planning benchmark. Through comprehensive experiments, we report new insights on the performance of representative path planning algorithms on planetary terrains, for the first time to the best of our knowledge. Our results show that classical algorithms can achieve up to 100\% global path planning success rates on average across challenging terrains such as Moon's north and south poles. This suggests, for instance, why these algorithms are used in practice by NASA. Conversely, learning-based models, although showing promising results in less complex environments, still struggle to generalize to planetary domains. To serve as a starting point for fundamental path planning research, our code and datasets will be released at: \url{https://github.com/mchancan/PlanetaryPathBench}.
\end{abstract}


\end{frontmatter}

\section{Introduction}

Autonomous robotic exploration over planetary surfaces, which has largely relied on rover platforms, has been a critical component of national and international space agency' missions over the past decades. With the advent of new Lunar and Martian missions, there is a need for advancing single- and multi-agent autonomous navigation research. Key challenges include strict safety and reliability requirements, particularly with extremely limited computing resources. Thus, research on rover task planning is being actively pursued \citep{cadre}.

\begin{figure}[ht!]
\centering
\includegraphics[width=\linewidth]{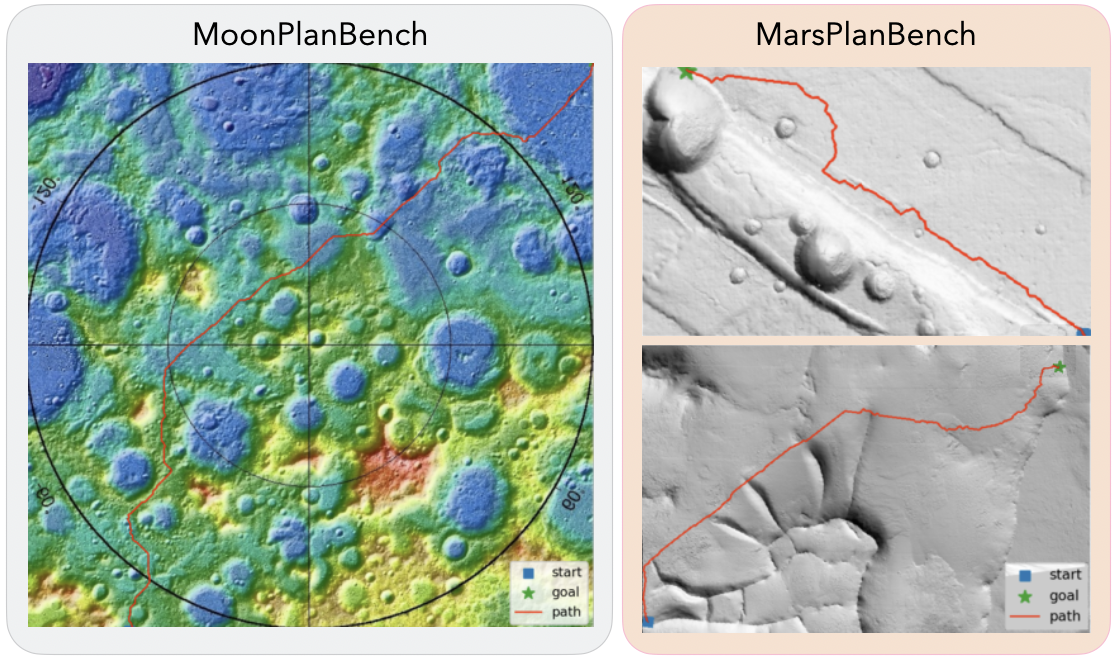}
\caption{We propose space mission-derived datasets of the Moon (left) and Mars (right), and report new insights on the performance of rover path planning algorithms.}
\label{f0}
\end{figure}

Despite decades of planetary exploration missions, there is still a lack of planetary datasets or benchmarks specifically designed for rover path planning and exploration. The vast majority of existing datasets are typically derived from planetary-analog terrains on Earth \citep{dlr, legged} or simulated environments with minimal planetary terrain fidelity such as synthetically generated \citep{omnilrs}. This has recently motivated the need for evaluating planning algorithms on planetary data. Although the data were derived from digital terrain models (DTM) or digital elevation models (DEM), most works are limited to single or few regions of interest \citep{vin, review-planetary}, visual semantic or grasping tasks \citep{luseg, lunar_grasp}, or other robotic platforms \citep{quadruped}, but not specifically for benchmarking rover global path planning.

Fundamental path planning research, on the other hand, focuses primarily on indoor or outdoor terrestrial environments \citep{vamp, gtmp}, with limited attention to planetary scenarios. The most similar attempt to using large amounts of planetary terrain data was by \cite{rover-irl}. However, they focus on path planning learning from human expert data by training an inverse reinforcement learning (RL) model, coupled with convolutional networks as in Value Iterative Networks (VIN), \cite{vin}, on a highly realistic synthetic path planning dataset. For this purpose, a heuristic tool developed by expert human rover drivers was used for model training to generate reference trajectories.

In this article, we take a step towards benchmarking global rover path planning algorithms on large amounts of data, derived from planetary missions, in a sustainable manner. Our goal is to build large planetary datasets and benchmarking tools for advancing fundamental research in classical and learning based path planning (see Fig. \ref{f0}). Our main contributions can be summarized as follows:

\begin{itemize}
\item The Mars Planning Benchmark (MarsPlanBench) dataset: a large planar dataset comprising over $2000$ 2D occupancy maps; derived from HiRISE DTMs of Mars, \cite{hirise}.
\item The Lunar Planning Benchmark (MoonPlanBench) dataset: over 36 2D occupancy grids of the Moon's north and south poles landing sites; derived from Lunar DEMs by \cite{lola}.
\item New insights on the performance of classical and learned global path planning algorithms for planetary exploration, consistent with recent research by \cite{enav}.
\end{itemize}

We conducted extensive experiments on our proposed planetary datasets and evaluated the most representative classical (graph, sampling) and learned (deep and RL) path planning algorithms. Our results show that classical algorithms, such as Dijkstra, still outperform learned models by substantial margins in terms of success rates, path lengths, and planning times. This highlights the robustness of these algorithms for global planetary planning tasks; validated in practice by NASA, \cite{enav}.

We believe that our large datasets and benchmarking framework will serve as a starting point for fundamental research on planetary rover path and motion planning, especially for advancing the state-of-the-art of learning-based methods. For the benefit of the community, we are releasing our benchmarking codebase and full datasets. And to enable sustainable growth of our approach, we are also making available our pre-processing code to generate more standardized datasets from planetary DTMs/DEMs.



\section{Preliminaries}

Here we describe a conventional formulation of the autonomous navigation and path planning problem for planetary rovers. Motion planning can be defined as the task of finding a rover motion states, from a start state to a goal state, while avoiding collisions with obstacles in an environment and satisfying several physical constraints such as joint, torque, or speed limits. We specifically focus on global path planning, a key component of motion planning, which is a purely geometric task for searching collision-free paths given a map of an environment. For rover explorers, these maps are typically planar occupancy grids.

\subsection{Global Path Planning}

The environment is represented as a 2D occupancy grid map $\mathcal{C} \in \{0,1\}^{W \times H} \subset \mathbb{R}^2$, in which a cell represents either free (0) space $\mathcal{C}_{free}$ or occupied (1) space $\mathcal{C}_{coll}$, where $W, H$ are the width and height of the grid, respectively. The state of a rover at any given time is defined by its position $\mathbf{p}_{t}= (x, y) \in \mathcal{R}^2$ within the grid. The objective is to find a collision-free geometric path configuration $\mathcal{P} = \{\mathbf{p}_1, \mathbf{p}_2, \ldots, \mathbf{p}_n\}$ from a start position $\mathbf{p}_0$ to a goal position $\mathbf{p}_n$ that minimizes a cost function $c(\mathcal{P})\in \mathcal{R}$. The cost function is generally subject to path length or traversal time optimization. Let $f:[0, 1] \to \mathcal{P}$, with $\mathbf{f}(t) \in \mathcal{C}$, to define its total variation as its arc length

\begin{equation} \label{eq:tv}
    TV(f) = \sup_{\mathcal{P}} \sum_{i=1}^{n} \| \mathbf{f}(t_i) - \mathbf{f}(t_{i-1}) \|.
\end{equation}

\textit{Definition 1 (Feasible Path)}: The function $f:[0, 1] \to \mathcal{P}$ with $TV(f)< \infty$ is 
\begin{itemize}
    \item a path, if it is continuous and of finite arc length,
    \item a feasible path, if and only if $\forall t \in [0, 1], f(t) \in \mathcal{C}_{free}$, $f(0)=\mathbf{p}_0$, $f(1) \in \mathcal{P}$.
\end{itemize}

Let $\mathcal{P}_{free}$ be the set of all feasible paths from $\mathbf{p}_0$ to $\mathbf{p}_n$ for a feasible path planning problem, without considering dynamic constraints, invalid states that violate collision constraints, or configuration limits, \citep{gtmp}.



\section{Methods and Benchmark Setup}

\subsection{Datasets}

Here we describe how MarsPlanBench and MoonPlanBench were derived. For preliminary validation experiments, we use the Radish dataset, widely adopted for evaluating robotic planning algorithms in large environments, as in prior work by \cite{gtmp}.

\subsubsection{MarsPlanBench.} Inspired on the Mars rover navigation experiments by \citep{vin}, we used the same source of DTMs from the Mars Reconnaissance Orbiter (MRO) mission \citep{hirise} to build MarsPlanBench. These DTMs have a spatial resolution of 1m or 2m per pixel and cover various Martian terrains, including plains, craters, and rocky regions. In VIN, the authors used a single DTM to generate 128$\times$128 image patches and obtain occupancy grids for training/testing purposes. For each map, a threshold of 10 degrees slopes or more was used to compute non-traversable cells.

We extend this idea by processing the entire collection of 1193 HiRISE DTMs, available as of Nov. $2025$. We process the entire data twice by thresholding elevation data to identify traversable and non-traversable areas, based on slope and roughness criteria relevant to rover navigation. This resulted in a total of $2386$ grid maps over two variants of the MarsPlanBench dataset, MarsPlanBench-10 and MarsPlanBench-20, with $10^\circ$ or higher and $20^\circ$ or higher slopes considered as non-traversable cells, respectively. The obtained occupancy maps have an average of $200\times400$ resolution, with a spatial resolution of $10$m per pixel on average. We note that we down-sampled all original DTMs images only for experimental and benchmarking purposes, given that the vast majority of path planning algorithms are typically evaluated on $2$D occupancy grids of such sizes. For our Mars experiments, we selected 306 challenging maps out of MarsPlanBench-10, which we are also releasing together with the full data. We also highlight that our code for pre-processing and generating occupancy maps, which is based on a codebase shared by \citep{vin}, is made publicly available and can be used to generate higher resolution maps accordingly.

\subsubsection{MoonPlanBench.} For building MoonPlanBench, we take inspiration from  \cite{review-planetary}. We derived $36$ planar occupancy grids from the Moon's north and south poles landing sites, from DEMs of the Lunar Orbiter Laser Altimeter (LOLA), an instrument onboard the NASA Lunar Reconnaissance Orbiter (LRO) mission, \cite{lola}. These DEMs have a spatial resolution of up to $5$m per pixel and capture unique topographical features of the Lunar poles, such as shadowed regions and rugged terrain.

We convert Lunar DEMs into occupancy grids using slope and roughness thresholds of $10^\circ$, $15^\circ$ and $20^\circ$ for rover traversability. We note that it was also necessary to down-sample original DEMs images by a factor of $64$ in order to obtain occupancy maps of manageable sizes for benchmarking. This resulted in an average of $400\times400$ grid cells per map, with varying spatial resolutions per pixel. We also release our code for generating higher resolutions occupancy maps from the original Lunar DEMs.

\subsubsection{The Radish Dataset.} We used a subset with the most challenging maps from the Radish dataset, following \cite{gtmp}. This resulted in $5$ occupancy grids collected from real-world indoor environments: Aces3, Freiburg, Intel Lab, Orebro, and Seattle. These maps have varying sizes and complexities, providing a diverse set of scenarios for evaluating path planning performance. All maps have grid sizes of less than $600\times600$ cells, with varying spatial resolutions per pixel.

\subsection{Classical Algorithms}

Classical path planning algorithms, typically built as a modular component of traditional navigation pipelines (e.g., localization, mapping, planning and control), have been widely used in robotics for decades. Therefore, there is a wide variety of implementations of some of the most representative path planning algorithms currently available. We therefore conducted a comprehensive search and evaluation of various open-source, functional implementations, and selected the most stable and performant ones to include in our benchmark. For benchmarking we consider planning algorithms from two main categories broadly used in the robotics community: graph-based search algorithms and sampling-based planners.

\subsubsection{Graph-based methods.} Graph-based methods, such as Dijkstra's algorithm and its variants AStart \citep{astar} and ThetaStar \citep{thetastar}, discretize the environment into a weighted graph structure, with non-negative edge weights, to find the shortest path from a start node to all other nodes. It uses a priority queue to explore nodes based on their cumulative cost from the start node. The shortest path is defined recursively as:

\begin{equation}
    d(v) = \min_{u \in N(v)} [d(u) + w(u,v)],
\end{equation}

where $d(v)$ is the shortest distance from the start node to node $v$, $N(v)$ is the set of neighboring nodes of $v$, and $w(u,v)$ is the weight (cost) of the edge between nodes $u$ and $v$.

\subsubsection{Sampling-based methods.} Sampling-based planners, such as Rapidly-exploring Random Trees (RRT) \citep{rrt} and its variants RRT Connect \citep{rrtconnect} and Dynamic RRT \citep{drrt}, randomly sample the map or configuration space $\mathcal{C}$ to build a tree of feasible paths. The RRT equation can be described as:

\begin{equation}
    \mathbf{p}_{new} = \mathbf{p}_{nearest} + \epsilon \frac{\mathbf{p}_{rand} - \mathbf{p}_{nearest}}{\|\mathbf{p}_{rand} - \mathbf{p}_{nearest}\|},
\end{equation}

where $\mathbf{p}_{new}$ is the new node added to the tree, $\mathbf{p}_{nearest}$ is the nearest node in the tree to the randomly sampled point $\mathbf{p}_{rand}$, and $\epsilon$ is a step size parameter that controls the distance between nodes.

In preliminary experiments, we tested other classical algorithms such as Probabilistic Roadmaps \citep{prm} and Artificial Potential Fields \citep{apf}, but they showed lower performance, in terms of path planning computation times, and were thus omitted from our main, large experiments for brevity. This was also due to the scale of our datasets, given that in our experiments we limit the planning time per trial, and these algorithms often require longer times to find feasible paths.

\subsection{End-to-end Learned Models}

Several machine learning-based models for path planning have been proposed and demonstrated in small, single planetary derived occupancy maps. We extend these evaluations and conduct preliminary experiments to characterize the performance of various learned models.

\begin{figure*}[ht]
\centering
\includegraphics[width=.32\linewidth]{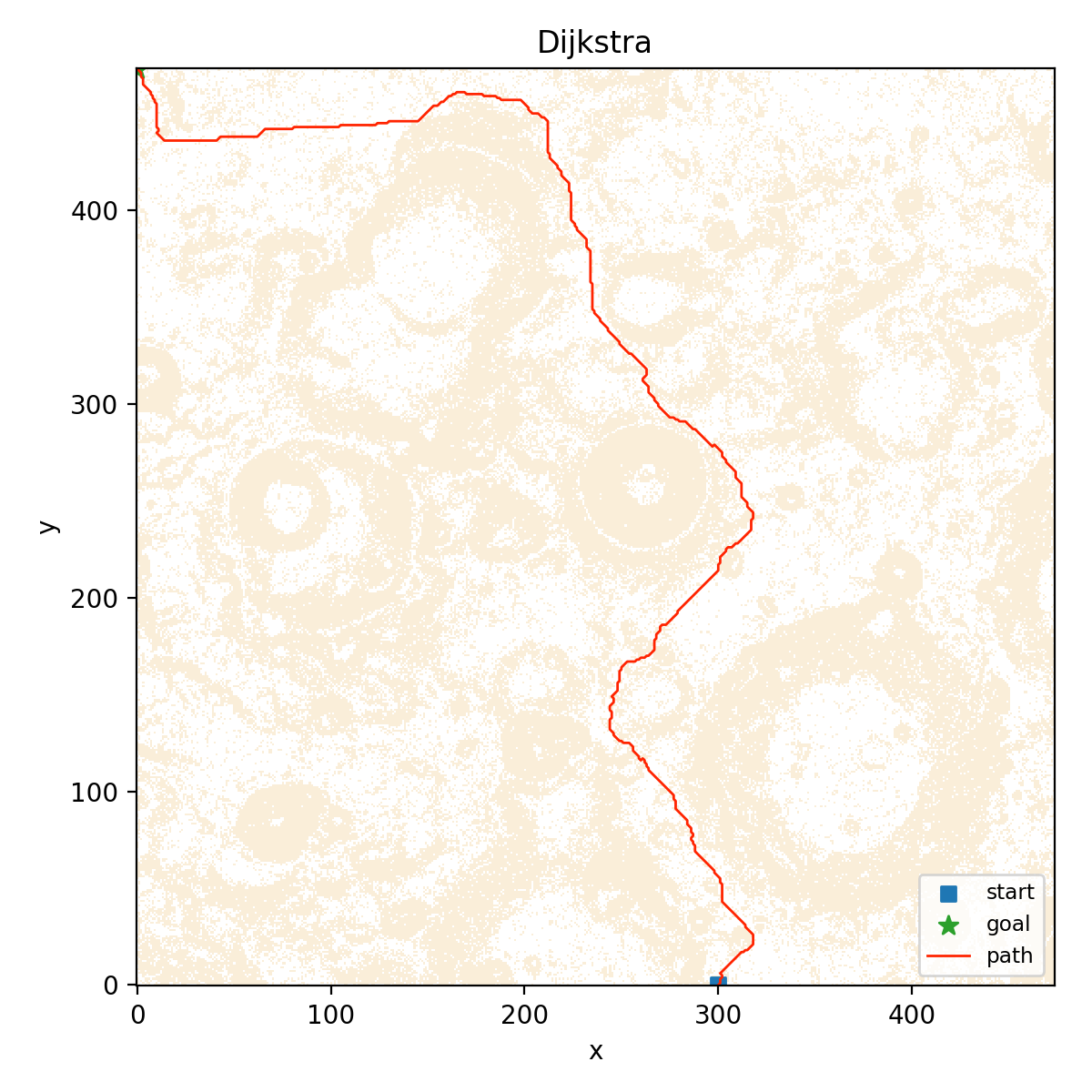}\hfill
\includegraphics[width=.32\linewidth]{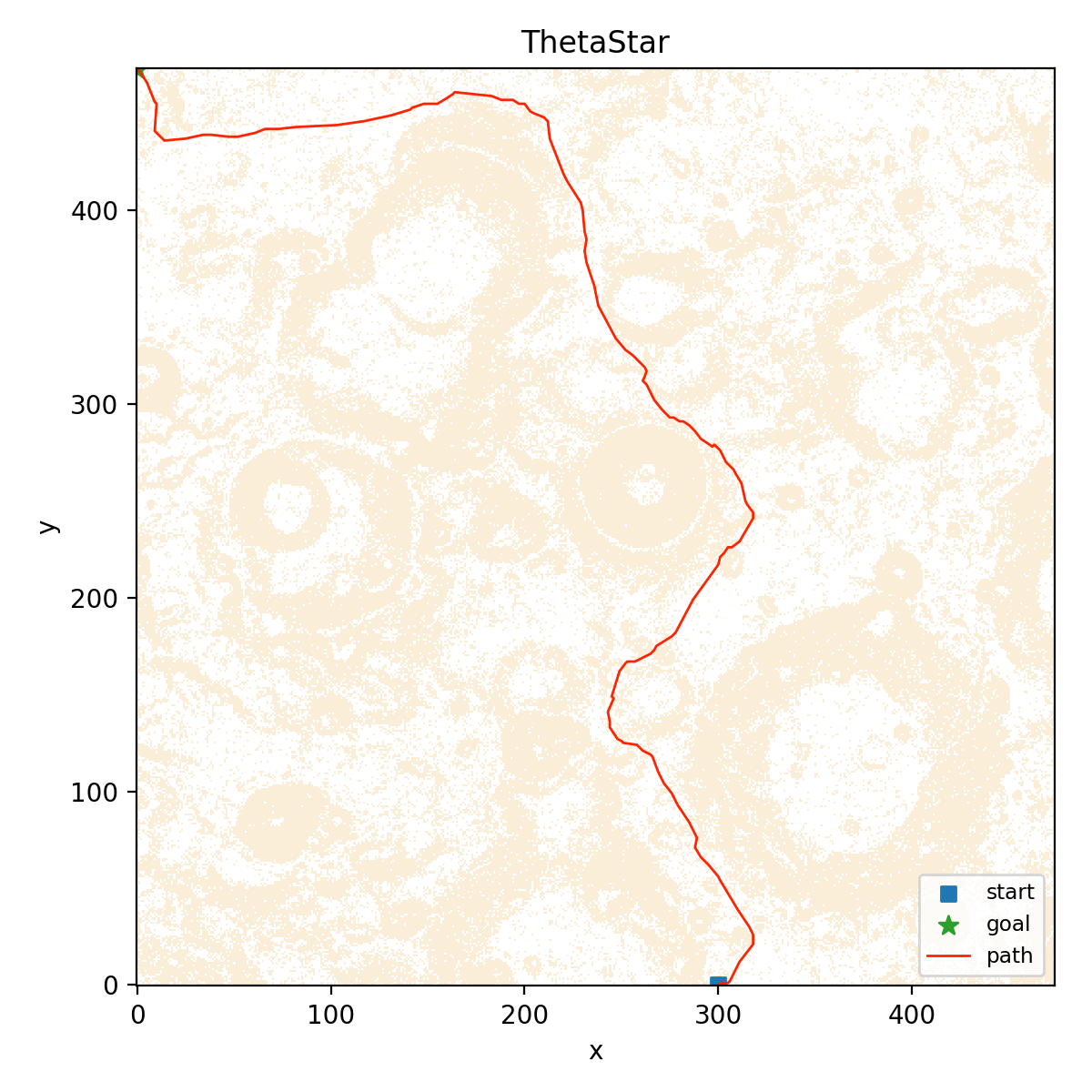}\hfill
\includegraphics[width=.32\linewidth]{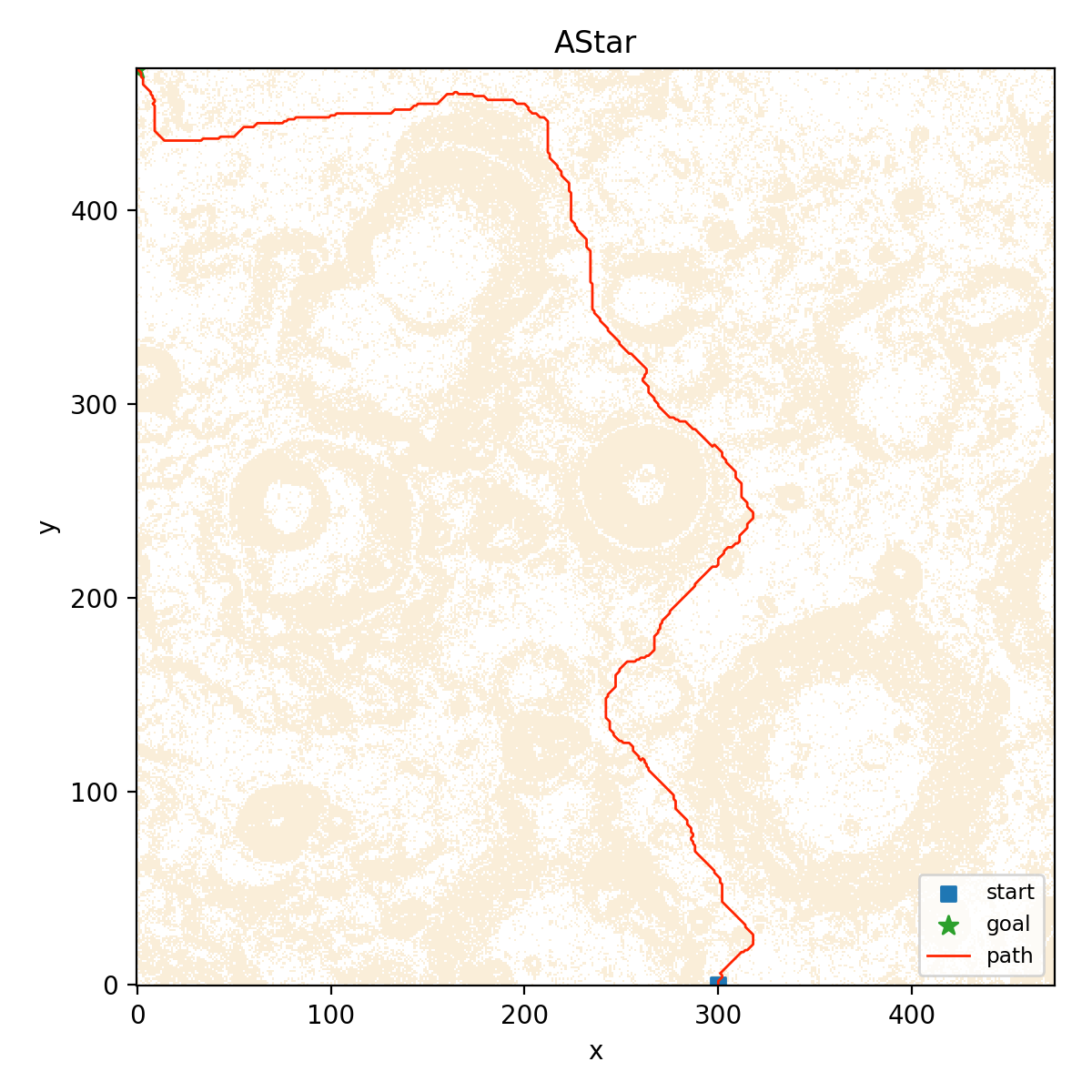}
\caption{Qualitative results of top-3 planners (Dijkstra, ThetaStar, AStar) on a map of the MoonPlanBench-10 dataset.} 
\label{f1}
\end{figure*}

\begin{table*}[ht] 
\centering
\caption{Averaged benchmarking results of top-6 path planning algorithms on all the variants of the MoonPlanBench dataset. Each dataset variant comprises 36 occupancy maps.}
\label{t1}
\begin{tabular}{lccccccccccccc}
 & \multicolumn{4}{c}{MoonPlanBench-10} & \multicolumn{4}{c}{MoonPlanBench-15} & \multicolumn{4}{c}{MoonPlanBench-20} \\
  \cmidrule(lr){2-5}
  \cmidrule(lr){6-9}
  \cmidrule(lr){10-13}
 & Success & Path & Plan. & Dist. & Success & Path & Plan. & Dist. & Success & Path & Plan. & Dist. \\
 & rate & lenght & time & left & rate & lenght & time & left  & rate & lenght & time & left \\
\hline
Dijkstra & 100 & \textbf{651.81} & \underline{23.31} & 0 & 100 & \textbf{636.16} & \underline{16.18} & 0 & 100 & \textbf{620.24} & 13.77 & 0 \\
ThetaStar & 100 & \textbf{654.81} & \underline{31.90} & 0 & 100 & \textbf{639.14} & \underline{23.26} & 0 & 100 & 623.17 & 12.36 & 0  \\
AStar & 91 & 639.94 & 30.82 & 56.1 & 83 & 621.25 & 32.45 & 72.1 & 100 & \textbf{620.24} & 19.52 & 0\\
RRT & 8 & 554.01 & 36.32 & 525.5 & 42 & 731.14 & 13.11 & 385.6 & 100 & 776.42 & \underline{6.13} & 0\\
Dynamic RRT & 8 & 589.91 & 13.49 & 525.5 & 50 & 744.75 & 21.14 & 333.8 & 100 & 737.53 & \underline{7.52} & 0\\
RRT Connect & 0 & 0 & 0 & 554.1 & 8 & 766.27 & 5.65 & 554.5 & 25 & 789.26 & 5.91 & 441.9\\
\hline
\multicolumn{13}{l}{\footnotesize \footnotesize{\textbf{Shortest} and \underline{fastest} planners with success rate (SR) of 100\%.}}
\end{tabular}
\end{table*}

One of the first learning-based planners proposed in the literature, based on a CNN architecture and reinforcement learning, is VIN \citep{vin}. VIN was demonstrated to successfully learn to plan paths on small grid-world environments, but also on Mars terrain maps from DTMs, \citep{hirise}. However, as the state-of-the-art progresses, other models have outperformed this framework, thus we evaluate these recent models here including Online LSTM and its variations such as Convolutional Auto-Encoder CAE-LSTM, Bagging LSTM, and Way Point Network (WPN) \citep{wpn}. WPN, in particular, integrates Online LSTM, CAE-LSTM, and Bagging LSTM.

In our experiments, we find that WPN outperforms all learned models on the Radish dataset with up to 100\% success rates on average. However, its main limitation was the path planning computing time, which was over 10$\times$ higher than classical methods. We note that when using space grade processors, which typically run at 200MHz, would further reduce algorithmic performances. We thus only include the results of WPN on the Radish benchmark for fairness, and show that all the other learned models were unable to find feasible paths on the Radish dataset. We note that our findings using learned models are also consistent with prior work, e.g., \cite{pathbench}, which highlights the challenges of these models to generalizing to unseen environments.

\subsection{Evaluation Metrics}

We evaluate the performance of each path planning algorithm using the following metrics, \cite{pathbench}.

\textit{Success Rate (SR)}. The percentage of maps, from a given dataset, in which a planner successfully finds a feasible path from start to goal within a specified time limit

\begin{equation}
    SR = \frac{N_{s}}{N_{m}} \times 100\%,
\end{equation}
where $N_{s}$ is the number of successful path planning tasks, and $N_{m}$ is the total number of maps in a dataset.

\textit{Planning Time ($T_{plan}$)}. The total computation time taken by a planner to find a feasible path, measured in seconds
\begin{equation}
    T_{plan} = t_{n} - t_{0},
\end{equation}

where $t_{0}$ and $t_{n}$ are the start and end timestamps of the planning process, respectively. The average planning time $\hat{T}_{plan}$ is computed across all maps in a dataset.

\textit{Path Length ($L(\mathcal{P})$)}. The total length of a planned path, calculated as the sum of Euclidean distances between consecutive waypoints along the grid cells

\begin{equation}
    L(\mathcal{P}) = \sum_{i=1}^{n} \| \mathbf{p}_i - \mathbf{p}_{i-1} \|,
\end{equation}

where $n$ is the number of waypoints in the path $\mathcal{P}$. The average path length $\overline{L}(\mathcal{P})$ is computed by averaging $L(\mathcal{P})$ across all maps in a dataset.

\textit{Distance Left ($D_{goal}$)}. The Euclidean distance from the last point of a planned path to the goal position. Typically used when a planner fails to find the goal
  
\begin{equation}
    D_{goal} = \| \mathbf{p}_n - \mathbf{p}_{goal} \|.
\end{equation}

where $\mathbf{p}_n$ is the last point of the planned path, and $\mathbf{p}_{goal}$ is the goal position. The average distance to goal $\overline{D}_{goal}$ is computed by averaging $D_{goal}$ across all maps in a dataset.

\textit{Path Deviation (PD)}. The path length difference when compared to a shortest possible path, here defined by the AStar algorithm

\begin{equation}
    PD = L(\mathcal{P}) - L_{A*}(\mathcal{P}),
\end{equation}

\begin{figure*}[ht!]
\centering
\includegraphics[width=\linewidth]{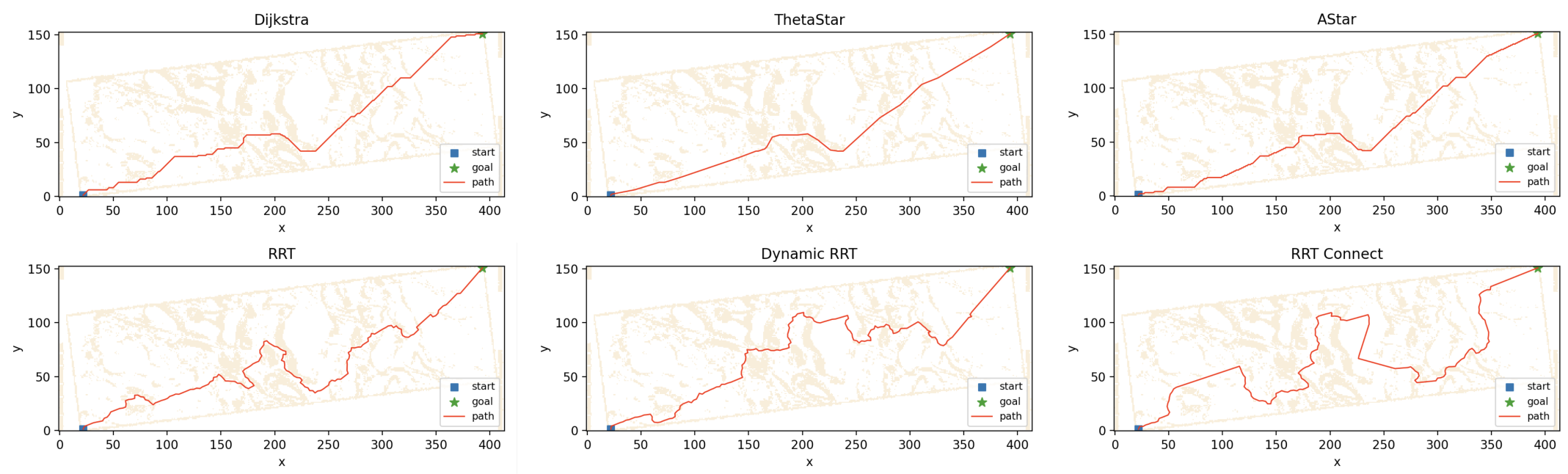}
\caption{Qualitative results of top-6 planners on a occupancy map of the MarsPlanBench-10 dataset.} 
\label{f2}
\end{figure*}

where $L_{A*}(\mathcal{P})$ is the length of the path planned by the AStar algorithm. The average path deviation is computed by averaging $PD$ across all maps in a dataset.

\textit{Memory Consumption (KiB)}. The total variation of memory used during a path generation task.

\textit{Trajectory smoothness (rad/move)}. The average change in heading direction per movement step on a planned path.

\textit{Obstacle Clearance (grid-cell units)}. The average distance from the planned path to the nearest obstacle.

\section{Experiments}

\subsection{Experimental Setup}

\subsubsection{Benchmarking Framework.} We build our unified framework for path planning evaluation on top of the PathBench library, \cite{pathbench}. First, we tested their implementations of classical and learned path planning methods on the Radish dataset, and then we incorporated other implementations we found performant and stable. This resulted in using the following implementations for our main experiments: Dijkstra, AStar, ThetaStar, RRT, Dynamic RRT, and RRT Connect from the Python Robotics repository, \cite{pythonrobotics}. And Online LSTM, CAE-LSTM, Bagging LSTM, and WPN from PathBench. All these implementations were carefully and systematically adapted to take occupancy grids as input.

\subsubsection{Path Planning Task.} To define start and goal locations across all maps, we created a function to strategically sample start and goal positions on each occupancy grid. This ensured that they are located in free space and are as far as possible from each other to require competent path planning capabilities. Along with our datasets, which includes png, csv and npy files, we also release the start and goal locations for each map to ensure reproducibility.

\subsubsection{Hardware and Evaluation Protocol.} We conduct extensive experiments on all three datasets using a commodity laptop computer with 13th Gen Intel Core i7-1355U, 32GiB RAM, running Ubuntu 24.04 with Python 3.8. We benchmark each planner with a maximum planning time of 60 seconds per map. The main goal of using this limit of time was allowing us to evaluate our entire datasets fairly quickly, with a focus on finding high-performance planners. For each dataset, we report averaged path planning metrics across all maps such as success rates in percentage (SR), path length in grid-cell units (Path Len.), planning time in seconds (Time), and distance to goal for each planner in grid-cell units (Dist. Left).

\begin{table}[t]
\centering
\caption{Averaged benchmarking results of top-6 planners on the selected 306 maps of the MarsPlanBench-10 dataset.}
\label{t2}
\begin{tabular}{lccccc}
 & Success & Path & Plan. & Dist. \\
 & rate & lenght & time & left\\
\hline
Dijkstra & 100 & \textbf{402.42} & \underline{3.1} & 0\\
ThetaStar & 100 & 402.82 & \underline{4.8} & 0\\
AStar & 100 & \textbf{402.42} & 6.5 & 0\\
RRT & 68 & 473.08 & 3.8 & 108.9\\
Dynamic RRT & 63 & 474.58 & 3.7 & 126.8\\
RRT Connect & 54 & 463.15 & 2.2 & 161.9\\
\hline
\multicolumn{5}{l}{\footnotesize \footnotesize{\textbf{Shortest} and \underline{fastest} planners with success rate (SR) of 100\%.}}
\end{tabular}
\end{table}

\subsection{Results}

\subsubsection{MoonPlanBench.} Fig. \ref{f1} shows the paths generated by the top-3 high-performing planners when evaluated on the MoonPlanBench-10. Table \ref{t1} summarizes the main results of our experiments on the full MoonPlanBench dataset. We observe that classical graph-based search algorithms, particularly Dijkstra and ThetaStar, consistently achieve high success rates of $100$\% across all slope thresholds ($10^\circ
$, $15^\circ$, and $20^\circ$). They also provide the shortest path lengths while maintaining competitive planning times. On MoonPlanBench-20, e.g., considering planners with 100\% success rate, graph-based methods achieve path lenghts of around 620 compared to over 730 for sampling-based methods. As the terrain complexity increases, e.g., MoonPlanBench-10/20, only Dijkstra, ThetaStar reach $100$\% success rates. This means, the others do not reach the goal across all maps, thus their path lengths can be lower or slightly higher. Overall, sampling-based planners like RRT and its variants show lower success rates, especially at lower slope thresholds, but their performance increases as the complexity of the terrains decreases.

Overall, the evaluated sampling-based planners tend to produce longer paths with higher distances left to the goal when they fail. However, they demonstrate faster planning times than graph-based search algorithms, particularly RRT and Dynamic RRT, at higher slope thresholds.

\subsubsection{MarsPlanBench.} Table \ref{t2} summarizes averaged benchmarking results of the top-6 high-performing path planning algorithms on the MarsPlanBench-10 dataset. We highlight that Dijkstra provides the shortest path on average with the quickest planning time, over $2\times$ faster than the top-2 planner. Fig. \ref{f2} illustrates the paths generated by the top-6 high-performing planners on a representative map from the MarsPlanBench-10 dataset.

\subsubsection{Radish.}  
\begin{figure}[t]
\centering
\includegraphics[width=\linewidth]{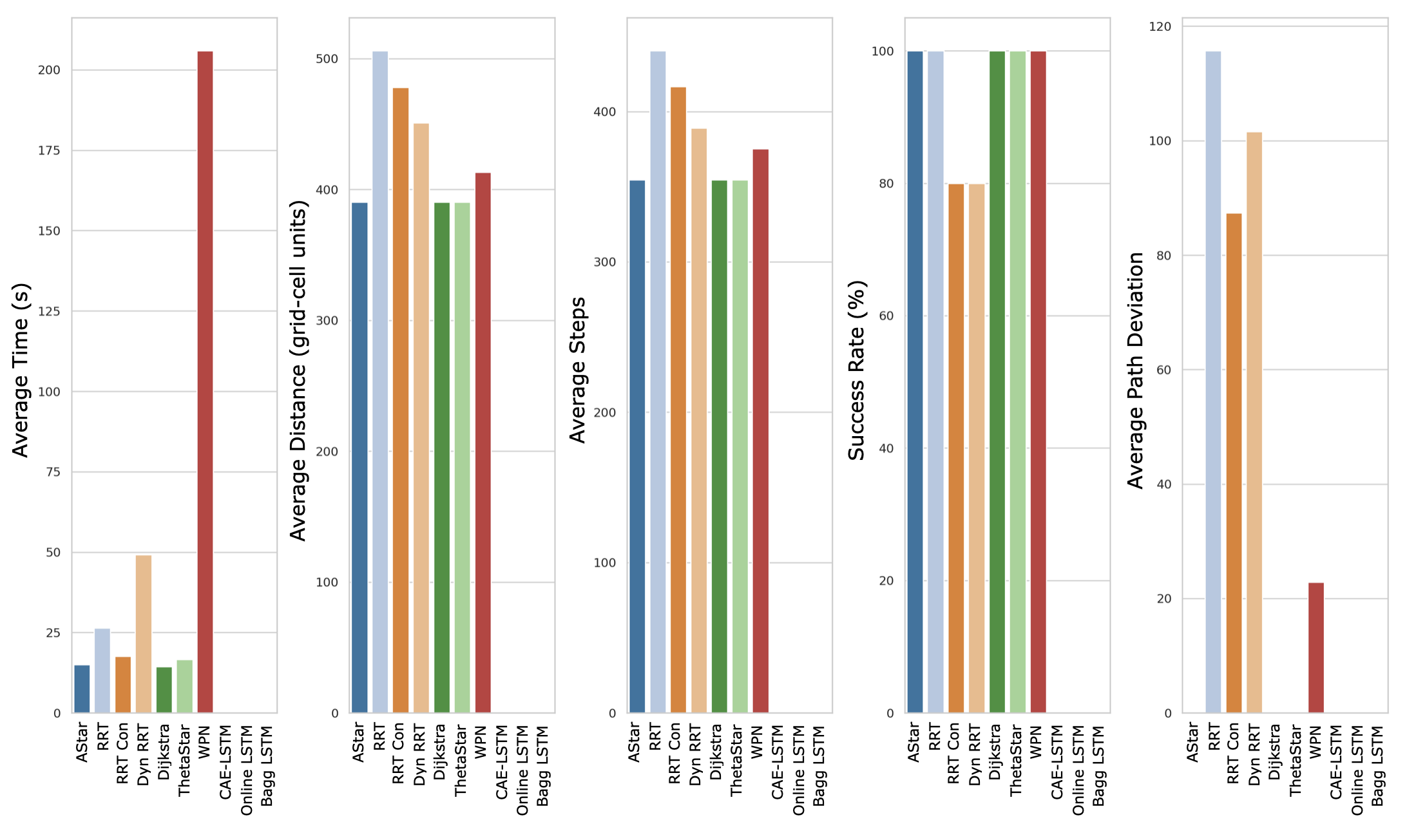}
\caption{Averaged results on the Radish dataset.} 
\label{f4}
\end{figure}

In Fig. \ref{f4} we report metrics such as average time, average distance, average steps, success rate, and average path deviation on the Radish dataset. We note that most learned models were unable to find feasible paths, except for WPN, which achieved an average success rate of 100\% but with approx. $10 \times$ higher planning times compared to classical methods. This highlights the challenges faced by learned models in generalizing to unseen environments, particularly in complex planetary terrains.

\begin{figure}[t]
\centering
\includegraphics[width=\linewidth]{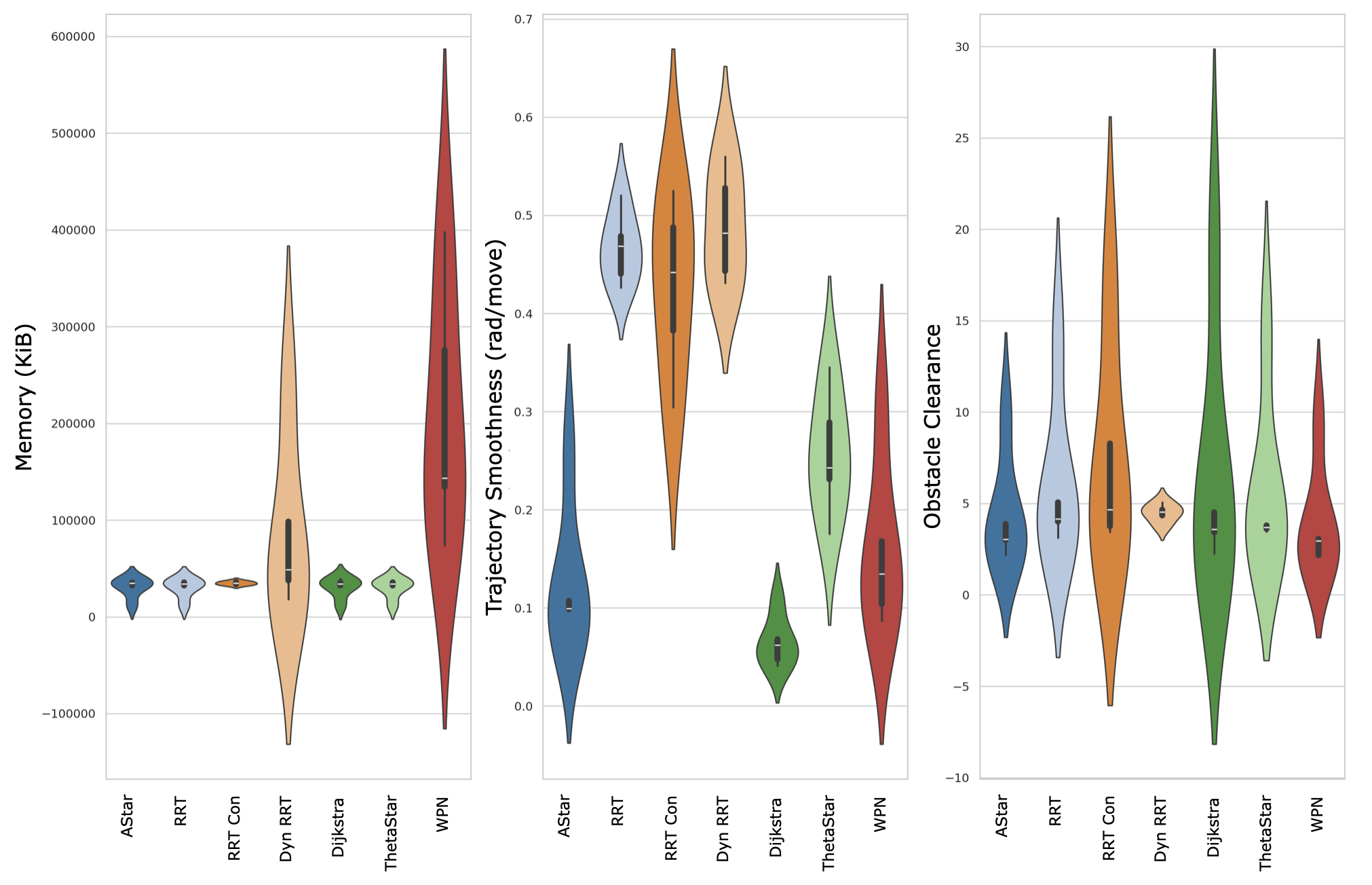}
\caption{Statistical results on the Radish dataset.} 
\label{f6}
\end{figure}

In Fig. \ref{f6} we analyze RAM usage, trajectory smoothness, and obstacle clearance on the Radish dataset. Dynamic RRT and WPN show high variance of memory consumption compared to all the rest of the planners. In terms of trajectory smoothness, Dijkstra produces the smoothest paths, while RRT variants , AStar and WPN tend to generate more erratic trajectories. Obstacle clearance is generally consistent across all planners, with much less variation for Dynamic RRT.

\subsection{Additional Qualitative Results}

\begin{figure}[b]
\centering
\includegraphics[width=\linewidth]{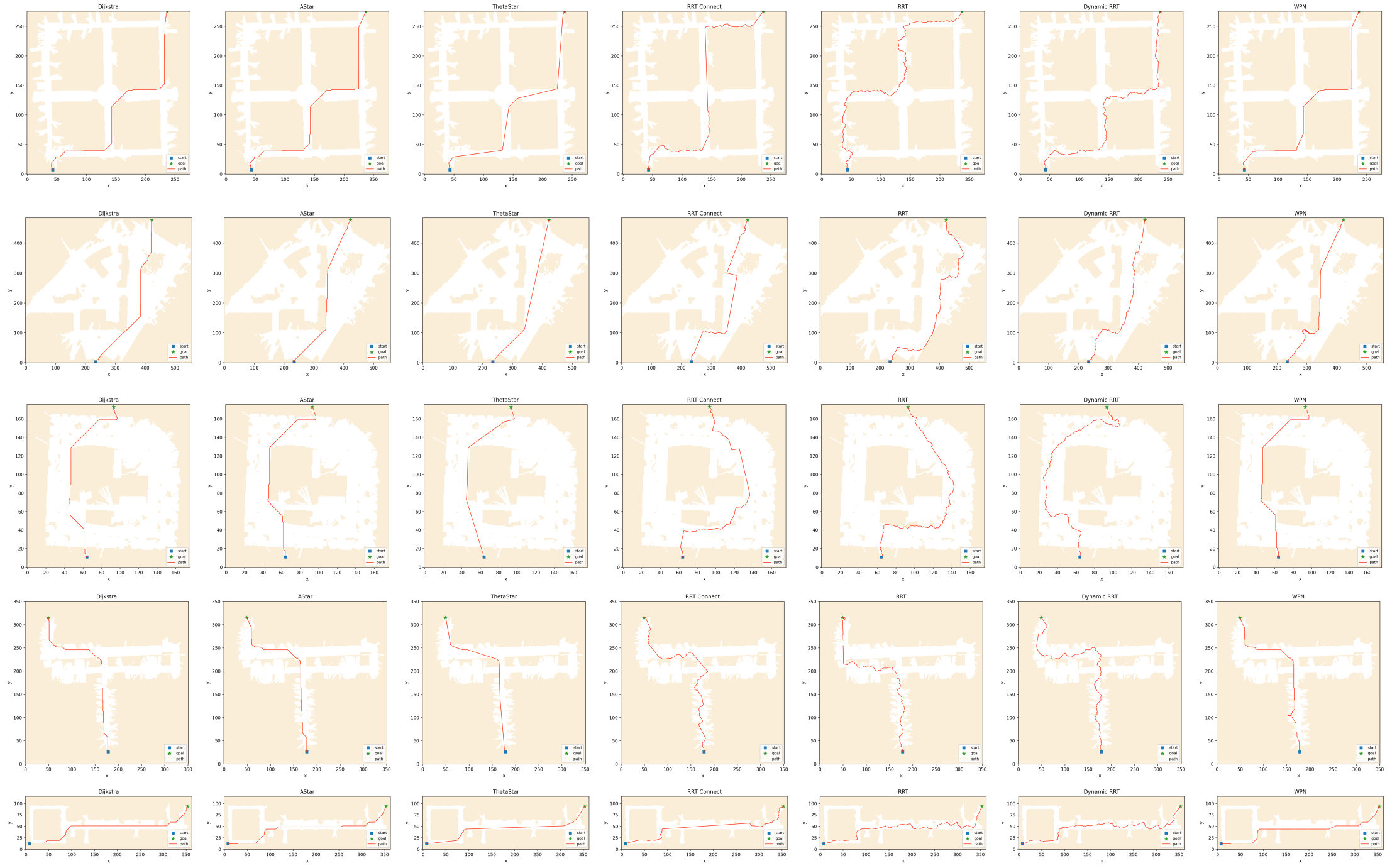}
\caption{Qualitative results of classic and learned path planners on the Radish dataset: (top row to bottom row) Aces3, Freiburg, Intel Lab, Orebro, and Seattle maps; (left to right) Dijkstra, AStar, ThetaStar, RRT Connect, RRT, Dynamic RRT, and WPN planners.} 
\label{f7}
\end{figure}

\begin{figure*}
\centering
\includegraphics[width=\linewidth]{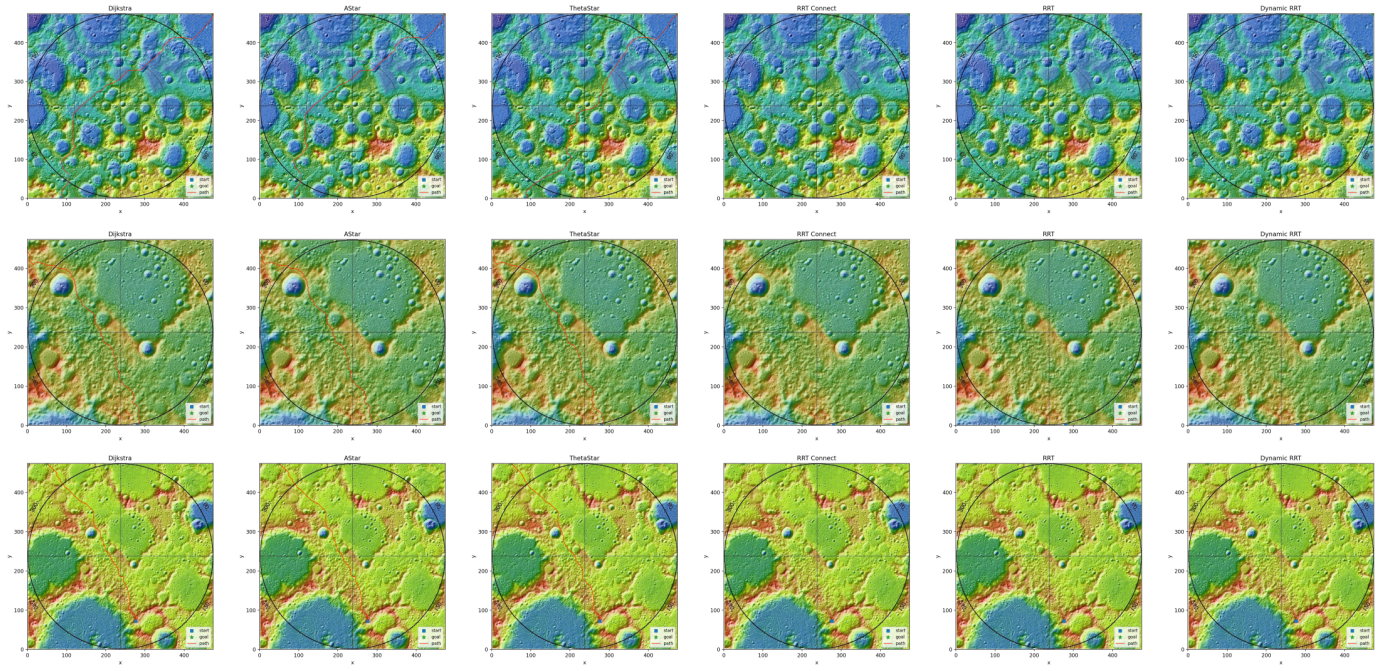}
\caption{Qualitative results on a selected subset of the MoonPlanBench-10 dataset; (left to right) Dijkstra, AStar, ThetaStar, RRT Connect, RRT, and Dynamic RRT planners. We overlay the planned paths on top of the original Lunar DEMs \citep{lola} for better visualization.}
\label{f8}
\end{figure*}

\begin{figure*}
\centering
\includegraphics[width=\linewidth]{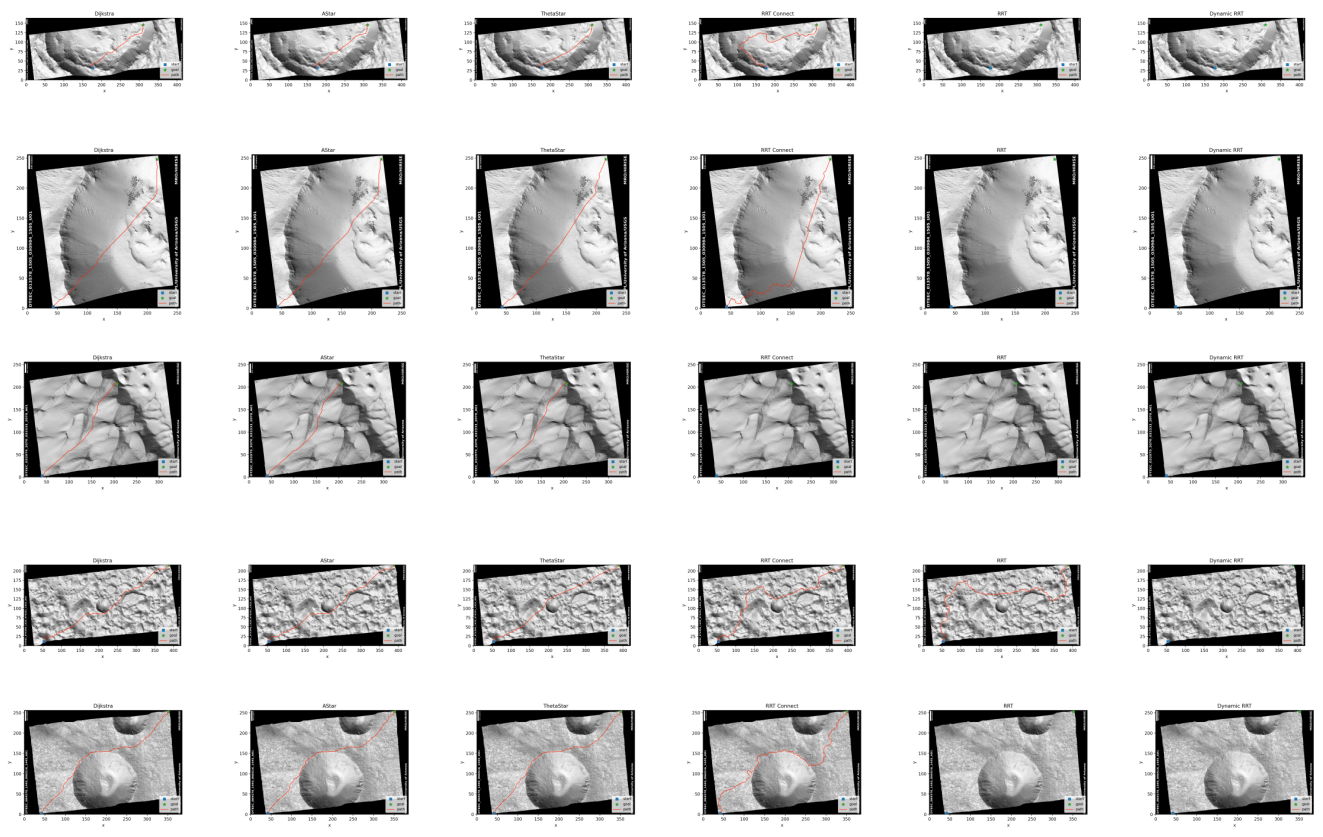}
\caption{Qualitative results on a selected subset of the MarsPlanBench-10 dataset; (left to right) Dijkstra, AStar, ThetaStar, RRT Connect, RRT, and Dynamic RRT planners. We overlay the planned paths on top of the original HiRISE DTMs \citep{hirise} for better visualization.}
\label{f9}
\end{figure*}

Figs. \ref{f7}, \ref{f8} and \ref{f9} show paths generated by classical and learned path planners on the Radish, MoonPlanBench, and MarsPlanBench datasets. As discussed in our main experiments, classical planners such as Dijkstra, AStar, and ThetaStar successfully find feasible paths across all maps, demonstrating their robustness in real-world indoor environments. For the learned models, only WPN manages to generate feasible paths on the Radish dataset.

\section{Conclusion}

Our main findings suggest that classical path planning algorithms, such as Dijkstra, demonstrate robust performance on planetary terrains. The Dijkstra algorithm achieved high success rates, efficient path lengths, and minimal planning time compared to other classical and learning-based methods across all our datasets. This can also be validated by a recent paper from the Jet Propulsion Laboratory \citep{enav}, where Dijkstra is used as the global path planning method in the autonomous driving algorithm of the Mars rover Perseverance.

A key limitation of using, for instance, HiRISE orbital images from Mars missions is that they might not be sufficiently detailed for fine grained terrain, exploration analysis due to their resolution constraints (max. res. being 5m per pixel). This is why we mainly focus here on global path planning tasks, and highlight that our key objective is to provide planetary datasets and benchmarks for enhancing fundamental autonomous navigation research.

Other algorithms such as D* \citep{dstar} and its variants were not included in our benchmark since they are primarily designed for dynamic replanning scenarios, which is outside the scope of this work. However, we consider Dynamic RRT, in our main experiments given its performance. Nevertheless, our codebase is capable of supporting new algorithms to allow much broader comparisons.

This work lays a foundation for more extensive research in planetary rover path planning. By releasing our codebase for preprocessing and generating standardized 2D occupancy maps, we enable the provision to expand the datasets using additional DEMs and DTMs from upcoming Lunar and Martian missions. The benchmarking framework presented here also paves the way for more focused research, highlighting key directions such as extending to 3D terrain representations with interactive simulations, incorporating motion planning with rover dynamics, and multi-agent strategies for cooperative exploration.


\begin{ack}
This work was partially supported by Kempestiftelserna (The Kempe Foundations), the Wallenberg AI, Autonomous Systems and Software Program (WASP) funded by the Knut and Alice Wallenberg Foundation, and the Swedish National Space Agency (SNSA). 
\end{ack}


\bibliography{ifacconf}             


\end{document}